\def\year{2021}\relax
\documentclass[letterpaper]{article} 
\usepackage{aaai21}  
\usepackage{times}  
\usepackage{helvet} 
\usepackage{courier}  
\usepackage[hyphens]{url}  
\usepackage{graphicx} 
\usepackage{graphicx}  
\usepackage{natbib}  
\usepackage{caption} 
\usepackage{todonotes}
\newtheorem{definition}{Definition}
\usepackage{amsmath}
\usepackage{bm}
\usepackage{bbm}
\DeclareMathOperator*{\argmax}{arg\,max}
\usepackage{multicol}
\usepackage{pdfpages}

\usepackage{subcaption}

\usepackage[section]{placeins}

\usepackage{cleveref}
\crefname{algocf}{alg.}{algs.}
\Crefname{algocf}{Algorithm}{Algorithms}

\newcommand{\norm}[1]{\lvert #1 \rvert}

\title{DocuBot : Generating financial reports using natural language interactions}
\author{

    Vineeth Ravi, Selim Amrouni, Andrea Stefanucci, Armineh Nourbakhsh, Prashant Reddy, Manuela Veloso \\

}
\affiliations{

    J.P. Morgan AI Research\\

    383 Madison Ave, Floor 21\\
    New York, New York 10036\\
    \{vineeth.ravi, selim.amrouni, andrea.stefanucci, armineh.nourbakhsh, prashant.reddy, manuela.veloso\}@jpmchase.com

}

\begin{document}

\maketitle

\begin{abstract}
The financial services industry perpetually processes an overwhelming amount of complex data. Digital reports are often created based on tedious manual analysis as well as visualization of the underlying trends and characteristics of data. 
Often, the accruing costs of human computation errors in creating these reports are very high. We present \textit{DocuBot}, a novel AI-powered virtual assistant for creating and modifying content in digital documents by modeling natural language interactions as ``skills'' and using them to transform underlying data. DocuBot has the ability to agglomerate saved skills for reuse, enabling humans to automatically generate recurrent reports. DocuBot also has the capability to continuously learn domain-specific and user-specific vocabulary by interacting with the user. We present evidence that DocuBot adds value to the financial industry and demonstrate its impact with experiments involving real and simulated users tasked with creating PowerPoint presentations.
\end{abstract}

\section{Background}
As part of their daily business operations, employees in the financial services industry often process large numerical data-sets and generate a variety of recurrent reports, including PowerPoint decks. The reports allow the employees to transform complex financial data such as cash-flows, client transactions, stock prices, market risk conditions, etc. into accessible visualizations and presentations. The data is often in time-series format, and many reports have a relatively consistent structure, with a need for recurrent updates on an annual, quarterly, monthly, weekly, or daily basis. The data can be overwhelming in the financial services industry, with companies like JPMorgan Chase processing more than $\$$1 Trillion in year-to-date volume for merchant clients in 2016 \citep{JPMorganPayment}, and generating upwards of 8 million PowerPoint slides every year\footnote{Source: internal J.P. Morgan study.}. As a result the report-generation process is often tedious, time-consuming, error-prone, and subject to complicated and costly controls.

Recent advances in data-to-text generation have proven promising, but human-level performance remains a challenge  \cite{shen2020neural, gatt2017survey, chen2020kgpt}. Furthermore, studies in this domain often focus on open-ended relational data rather than numerical data, and the output is often text-only commentary paired with no visualizations. For the use cases specific to the financial domain, commercial solutions such as \textit{Narrative Science} and \textit{Yseop} have been developed. These solutions allow users to set up sophisticated configurations that can be reused to generate new reports based on pre-existing templates. While configurability is helpful, a more flexible solution---one which could learn from direct interactions with end users---would be more preferable, easier to scale, and less challenging to generalize to new use cases.

In this paper we introduce \textit{DocuBot}, a novel framework to automate the generation of digital reports---specifically PowerPoint slides---in a real-world setting through human-AI interaction. DocuBot provides the ability to create and modify content in PowerPoint presentations through natural language instructions, with the capability to adapt and improve by learning from experience through interactions with the user. DocuBot also has the capability to automatically generate ``insights'' which are natural language explanations of data and content displayed on the slides in these presentations. Users can utilize this framework and create digital presentations effortlessly, by issuing natural-language instructions to DocuBot, similar to a conversational virtual assistant. However despite many other virtual assistants, DocuBot's tasks are more structured and targeted. This motivates the use of automation methods that can leverage business logic and structural constraints, while also maintaining flexibility and expressivity. In this regard, DocuBot's novel innovations include:
\begin{itemize}
    \item A simple CRF-based parser that identifies major concepts in user commands and makes sense of them within the boundaries of a what a typical business user would need. 
    \item An adaptive Knowledge Base that learns domain-specific or user-specific lingo via dynamic learning-and-forgetting mechanisms.
    \item A simulated user study that models synthetic user agents with varying degrees of predictability, cooperativeness, and diversity in their vocabulary. The study allows us to benchmark the performance of two Knowledge Representation approaches against each other and measure the impact of various user behaviors on DocuBot's performance. 
    \item An end-to-end pipeline designed along the common business need for automated generation of reports and presentations based on financial data. 
\end{itemize}

The following section describes DocuBot's architecture and each component in detail. The remaining sections describes how each component is evaluated or stress tested. We also present illustrations of user interactions with DocuBot and its output, and qualitative user feedback on the time saved by using DocuBot to generate PowerPoint slide decks.

\section{Components}

\begin{figure}[!htb]
\centering
\includegraphics[width=0.47\textwidth]{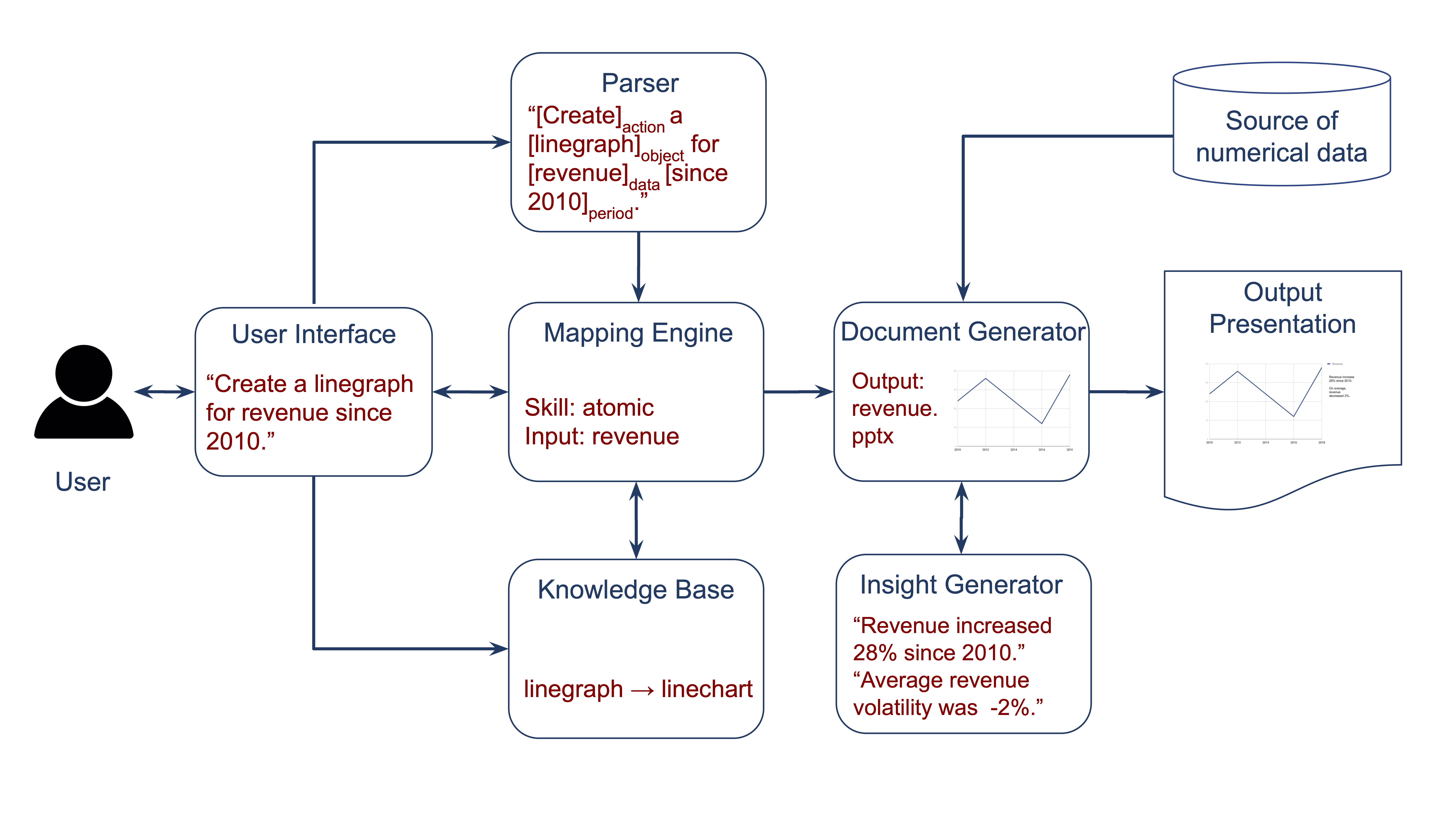}
\caption{DocuBot's architecture and components.}
\label{fig:experiment}
\end{figure}

In this section, we present how DocuBot is structured and describe each component in detail. 



\subsection{User Interface}

Any user-facing AI assistant needs a way to communicate with users, receive commands, ask clarifying questions, and produce proper answers. The user interacts with DocuBot through natural language instructions to create as well as modify digital presentations. The interface enables DocuBot to clarify any existing ambiguity present in the instructions as well as learn domain-specific vocabulary over time through interactions with users. 

In a later section, we demonstrate through screenshots of a working demo how a user interacts with DocuBot to create as well as modify content for PowerPoint slides.

\subsection{Parser}\label{sec:parser}

  To understand and execute requests, DocuBot requires an intent-aware parser that is able to process and transform user commands into relevant actions. Since DocuBot targets a specific task, the intents can be restricted to a closed set like \textit{updating client A's slides using client B's data}, or \textit{generating a pie chart using 2019 data in `data.csv'}. Working backwards from the intents allows  to identify the main pieces of information that DocuBot will need to perform each intended task. These pieces, known as ``concepts'', include\footnote{The deployed version of DocuBot covers multiple additional concepts such as \textit{client name}, \textit{period}, and \textit{type of analysis}. These concepts have not been described in the section because they were not included in the experiments at the time of writing this paper.}:
 \begin{itemize}
 \item \textit{action}: such as creating or updating a deck.
 \item \textit{data}: the dataset to be used as the source.
 \item \textit{object}: the type of graphic to be generated such as a piechart.
 \item \textit{presentation}: the name of the resulting presentation. 
 \end{itemize}
 
 With the above schema, the task of parsing user commands is reduced to identifying the above concepts in each command, to understand human instructions. In other words the parser needs to transform ``Please create a Piechart using Energy data and add it in the weekly report.'' into ``Please $[create]_{action} $ a $[Piechart]_{object} $ using $[Energy]_{data} $ data and add it in $[weekly report]_{presentation}$ .''
 
This can be treated as a tagging problem with relatively low lexical diversity (because there are not many ways in which a piechart can be mentioned). Our solution uses a CRF tagger trained on 50 natural language commands. The commands were curated from users across J.P. Morgan and annotated manually for concepts. We used the NLTK library \citep{Bird2009book} to assign Part-of-Speech tags to the commands. Each token was represented as a feature vector by concatenating the features in Table \ref{tab:features}. Next, we trained a Conditional Random Field \citep{Lafferty2001crf,Sha2003CRFparse,Sutton2012crf} model as implemented in CRFsuite \citep{CRFsuite}, and called through the python-crfsuite package \citep{python-crfsuite}. Despite the small size of the training data, the resulting model (evaluated on a test set of 25 unseen commands), performed relatively well (Macro-averaged F1-Score of 0.849, Precision of 0.86, Recall of 0.84). Hence we did not deem necessary to curate more data. The Parser's performance continued to prove robust during user testing.

\begin{table}[]
\caption{High-level features used by the parser to identify concepts in each command.}
\label{tab:features}
\begin{tabular}{ll}
\textbf{Features}                                                                    & \textbf{\begin{tabular}[c]{@{}l@{}}Example for\\ ``piechart''\end{tabular}} \\ \hline
Part-of-Speech                                                                       & Noun                                                                      \\ \hline
\begin{tabular}[c]{@{}l@{}}POS tag of previous\\ and next word\end{tabular}          & \begin{tabular}[c]{@{}l@{}}Proposition,\\ Verb\end{tabular}               \\ \hline
First and last letters                                                               & ``p'', ``t''                                                                      \\ \hline
\begin{tabular}[c]{@{}l@{}}Word with truncated\\ first and last letters\end{tabular} & ``iechart'', ``piechar''                                                          \\ \hline
\begin{tabular}[c]{@{}l@{}}2- and 3-letter\\ subwords\end{tabular}                   & \begin{tabular}[c]{@{}l@{}}``pie'', ``char'', ``t'', ``pi'', \\``ec'', ``ha'', ``rt'' \end{tabular}                                            
\end{tabular}
\end{table}

\subsection{Knowledge Base}
The vocabulary commonly used by users can be inconsistent. For example \emph{``J$\&$J''} is commonly used when communicating with DocuBot to create PowerPoint slides about the company \emph{``Johnson and Johnson''}. Furthermore, the vocabulary can be context-specific. For example depending on each user's intent, \emph{``graph''} could mean either \emph{``Piechart''} or \emph{``Histogram''}. It is difficult to have a consistent and exhaustive vocabulary mapping across all users in a large firm. DocuBot overcomes this limitation by having the ability to dynamically adapt and improve its performance through interactions with users for feedback, learning from experience.

Similar to NELL \citep{NELL-aaai15}, DocuBot can learn from experience by interacting with the user. It employs a continuously learning ``Knowledge Base'' (KB), which enables a dynamic mapping between natural language input and skills. Given the concepts tagged by the Parser, the KB models them into a hierarchy that can evolve over time for each user. The hierarchy includes \emph{main-concepts}(MC), \emph{sub-concepts}(SC), and \emph{vocabulary}. For instance the main-concept $chart$ can refer to a set of sub-concepts $piechart$, $barchart$, etc. A sub-concept such as $piechart$ can itself be expressed using a diverse vocabulary such as $pie$, $piegraph$ or $pizzachart$. The sub-concept $barchart$ can be expressed as $histogram$ or $barplot$. The KB maintains and updates mappings between main-concepts and sub-concepts as well as sub-concepts and vocabulary.


 We present two possible configurations for the KB, as described in the below sub-sections. To maintain consistency, the following notation holds throughout the paper:
 \begin{definition}
 \mbox{}
 \begin{itemize}
 \item $\bm{\mu}\in M$ refers to a main-concept (e.g. $chart$).
 \item $\bm{\sigma}\in S$ refers to a sub-concept (e.g. $piechart$, $barchart$).
 \item $w_c\in\bm{\omega}$ refers to a word in the user's vocabulary (e.g. $piegraph$, $histogram$). 
 \item The output of the Parser is a set of tuples in the form of (main-concept, word) or ($\mu$, $w_c$) $\in \Theta$.
 \end{itemize}
 \end{definition}
 

\subsubsection{Naive Knowledge Base}
\label{sec:NKB}
The Naive Knowledge Base (NKB) aims to learn the vocabulary employed by users referring to a concept by permanently mapping the first new word $w_c$ $\in$ $\bm{\omega}$ learned from user input. After obtaining the Parser output, the NKB enables the mapping from $\Theta$ to $S$. To do so, the NKB implements the following functionalities: (i) \emph{isInKB}: if $w_c$ exists anywhere in the KB accept, otherwise reject, (ii) \emph{inferSC:} return the $\bm{\sigma}$ linked to the $w_c$, and (iii) \emph{addToKB}: add the mapping from $w_c$ to $\bm{\sigma}$.

The learning mechanism is pretty simple. If a $w_c$ is rejected, DocuBot asks for clarification and uses the clarifying response to link it to the proper $\sigma$. For instance if the user asks for a ``pizzachart'' and DocuBot cannot find it in the NKB, it asks for clarification. If the user then asks for a ``piegraph'', the NKB may observe that the ``piegraph'' word has previously been linked to the ``piechart'' sub-concept, so it maps ``pizzagraph'' to the same sub-concept. In this mechanism learning is possible but there is no capacity for forgetting, therefore there is a risk of learning incorrect mappings.  


\subsubsection{Robust Knowledge Base}
\label{sec:rkb}
The NKB is limited in its flexibility and capacity to learn mappings of new vocabulary. The Robust Knowledge Base (RKB) is designed to overcome the limitations of the NKB. We introduce the notion of a $BeliefScore$ defined below:

\begin{equation}
B(\mu, \sigma, w_c) = Pr[w_c \in S].
\end{equation}
where $B(\mu, \sigma, w_c)$ is the $BeliefScore$ the Robust Knowledge Base has for a word ($w_c$) in a sub-concept ($\sigma$) belonging to a main-concept ($\mu$). 

 The RKB enables the mapping from $\Theta$ to $S$ and has ability to \emph{forget} incorrect mappings given by malicious users. Instead of permanently mapping the first input, RKB maintains a $BeliefScore$ of each triple $\mu$-$\sigma$-$w_c$. Every time a user interacts with DocuBot to choose a sub-concept, the RKB updates the discrete probability distribution and re-normalizes the score. The RKB implements the following functionalities: (i) \emph{isInKB}: if the triplet $\mu$-$\sigma$-$w_c$ is written in the KB accept, otherwise reject, (ii) \emph{addToKB}: add $\mu$-$\sigma$-$w_c$ to the RKB with an initial $BeliefScore$, (iii)  \emph{inferSC}: return the sub-concept that maximizes the belief score, or $\argmax_{s \in S}B(\mu, s, w_c)$, (iv) \emph{increaseBelief}: increase the $BeliefScore$ of the triplet $\mu$-$\sigma$-$w_c$, and (v) \emph{decreaseBelief}: decrease the $BeliefScore$ of the triplet $\mu$-$\sigma$-$w_c$. 
 
 If $l$ is the number of slides created so far, every time a new mapping is added to the RKB, we have the following formulas for updating the $BeliefScore$: 
 
 \begin{equation}\label{eq:ib}
 B(\mu, \sigma, w_c) \xleftarrow[]{\text{increaseBelief}} B(\mu, \sigma, w_c)*\frac{l-1}{l}+\frac{1}{l}
 \end{equation}
 Then
 \begin{equation}\label{eq:db}
 B(\mu, \sigma, w_c) \xleftarrow[]{\text{decreaseBelief}} B(\mu, \sigma, w_c)*\frac{l-1}{l}
 \end{equation}
 
 Eq \ref{eq:ib} and \ref{eq:db} ensure that a normalized score is maintained for each mapping. In later sections, we present experimental results on the NKB and RKB and their update mechanisms. 

\subsection{Mapping Engine}
While the Knowledge Base keeps track of terminology, the Mapping engine is an algorithm to map the concepts present in each natural language instruction into structured actions. The Mapping Engine enables DocuBot to jointly map the concepts identified in a command to one of possible action scenarios a.k.a \emph{skills} (described below). The mapping engine interacts with the Knowledge Base, Parser $\&$ the User-Interface to clarify any ambiguity in intent, so DocuBot can identify the corresponding available skills to use for appropriate content creation or modification of digital documents.

\subsubsection{Skills}
\label{sec:skills}
DocuBot has the flexibility to use various types of underlying data to automatically generate documents such as digital presentations like PowerPoints, PDFs, etc., and output files such as JSON requests, which are of great significance to business management and technology teams. DocuBot utilizes the predicted output concepts for mapping human instructions to corresponding \emph{skills} i.e. the various available content creation and modification capabilities. While DocuBot's framework is widely applicable to any general set of skills, in this paper for demonstration purposes we leverage the set of skills available in the python-pptx library \citep{python-pptx} for generating PowerPoint decks\footnote{These include line-charts, bar-charts, pie-charts, and other standard charting capabilities.}. However, internally at J.P. Morgan we have used the framework for non-standard visualizations such has water-fall charts, box-whisker plots etc., as well as diverse presentation formats. DocuBot's skills may vary for different business use-cases, and similar to human-learning, DocuBot's skills can be continuously enhanced and new skills could be added or learned through experience. In this paper we discuss three types of DocuBot's skills: (i) Atomic, (ii) Macro and (iii) Insights Generator Skills. The first two skills are described in the remainder of this section, while the last skills is described as a separate component in the following section.

\paragraph{Atomic Skills} refers to tasks that create or modify the contents of one slide or `object' in a digital presentation from a single natural language input command from the user.  The parameters of date and title in the slides as well as the location of data values in the data source files are auto-generated from templates used in  reports, which are common in business teams. Examples of natural language commands performing Atomic Skills: (i) ``Please [create]\textsubscript{action} a [Piechart]\textsubscript{object} about Share Performances using [market daily OHLC data set]\textsubscript{data} and include it in ['share performance report']\textsubscript{presentation} presentation.''.

\paragraph{Macro Skills} can create or modify the contents of multiple slides, `objects' or the entire digital presentation from a single natural language input command from the user, e.g. ``Please [create]\textsubscript{action} a [CompanyBriefingDeck]\textsubscript{object} using [Finance]\textsubscript{data} data and add it in [weeklyreport]\textsubscript{presentation} deck.'' Using this command, DocuBot creates a predetermined template named ``CompanyBriefingDeck'', of 10 slides using data in the ``Finance'' file, and adds the slides to a PowerPoint presentation named ``weeklyreport''.  The PowerPoint slides in \cref{fig:slide11}, \ref{fig:slide12}, \ref{fig:slide21}, and \ref{fig:slide22} are examples of DocuBot automatically generating and modifying content using Macro Skills. In the figures we show two slides out of the 10 slides deck([weeklyreport]\textsubscript{presentation}).

\paragraph{The Main Impact} is the re-usability enabled by this \emph{skillification}. Macro Skills become a simple aggregation of Atomic Skills and can be repeated over many recurrent updates. Another novel contribution of DocuBot is it's capability to save the previous commands of a user, encapsulating them as a new `object'. This allows it to  acquire new `skills' over time and enables end-users to re-use the new `skills' for future presentations.



\subsection{Insights Generator}
\label{sec:insightgenerator}
The previously described components help DocuBot understand and normalize human input. Once human input is properly mapped to concepts and skills, DocuBot needs to generate the visualizations and commentary necessary to be included in the output presentation. As previously described, the visualizations (e.g. piecharts, histograms, etc.) are created based on directions from the user. But presentations that only include visualizations with no context and no description are not useful. Each visualization needs to be accompanied by commentary that summarizes it and describes the most important and relevant aspects of its underlying data. We refer to this natural-language commentary as \emph{insight}. The Insights Generator component performs this task, using the following components:
(i) a set of \emph{primitives} that generate insights from the raw data, (ii) the mapping from insights to human-friendly text, (iii) a novel technique for the ranking and selection of most important insights, and (iv) a novel capability for hierarchical analysis of insights. 


\subsubsection{(i) Insight Generator Primitives}
We define \emph{primitives} as the set of the different numerical operations that can be applied to the underlying numerical data which is modeled as a time-series.  Examples of primitives include: (1) Absolute value primitives: These primitives compute metrics on the raw value of the time series: minimum, maximum, rolling average, volatility, etc. Or, (2) Comparison primitives: These use access to the full history of the data to compute metrics about the time series and then compare the value at any slice: distance to the mean, \emph{comparative factor} \citep{Perera2018thesis}, etc. The set of primitives is expected to grow over time.



\subsubsection{(ii) Text Generation from Insights}
Data-to-text generation is a growing area of research in NLU and NLG \citep{shen2020neural, gatt2017survey}. However performance is still far from human baselines, which is a major hurdle to deployment in enterprise settings. Furthermore, user surveys revealed that users were interested in having consistent, well defined, and reliable text output. Therefore we opted for a template-based approach. Relevant templates were curated as part of DocuBot's development process, which involved extensive engagements with end-users. The templates were implemented as a sequence of interchangeable slots, e.g. ``$<$company share$>$ averaged $<$rate$>$\% daily return.'' 
\footnote{We have since experimented with PCFG-based approaches which will be described in upcoming publications}.

This process was also useful in collecting new primitives that were relevant for final users. For example, in a particular business use-case DocuBot was required to generate insights for a new Key Performance Indicator (KPI). DocuBot was beneficial to the firm's executive management by highlighting relevant facts about KPIs amidst disruptive events such as the COVID-19 pandemic. 


\subsubsection{(iii) Insights Ranking and Selection}
 Not all possible commentary is interesting or relevant to users. Additionally, in certain cases the commentary can be overwhelming. In one case prior to the implementation of this module, DocuBot was generating 12,000 daily insights. 

DocuBot aims to provide the user with only the most interesting and valuable insights. For that purpose, it needs to rank and/or select insights. In order to sort the insights, we define their importance by a set of utility scoring functions such as \textit{impact on internal revenue}, \textit{anomaly compared to previous period}, and \textit{anomaly compared to peers}. Each insight is ranked by an interpolated aggregate of all utility weights, and the top $K$ insights are displayed. The interpolation weights and $K$ are all configurable by end-users. 


\subsection{Document Generator}
As previously mentioned, for demonstrating DocuBot's capabilities in this paper, we use the python-pptx library \citep{python-pptx} to generate presentations. The deployed version of the tool supports multiple other outputs such as PDFs, Microsoft Word documents, and web-pages. 


\section{Experimental Results and Discussion}



In this section, we present experiments and discuss DocuBot's Knowledge Base. Further, we demonstrate the advantages of employing the Robust Knowledge Base (RKB) over the Naive Knowledge Base (NKB) in the presence of Non-Collaborative Users. We show that the RKB learns new vocabulary faster and has superior performance in mapping user intent to output formats consistently across all experiments.


\subsection{Naive Knowledge Base (NKB)}

\begin{figure}[!htb]
\centering
\includegraphics[width=0.5\textwidth]{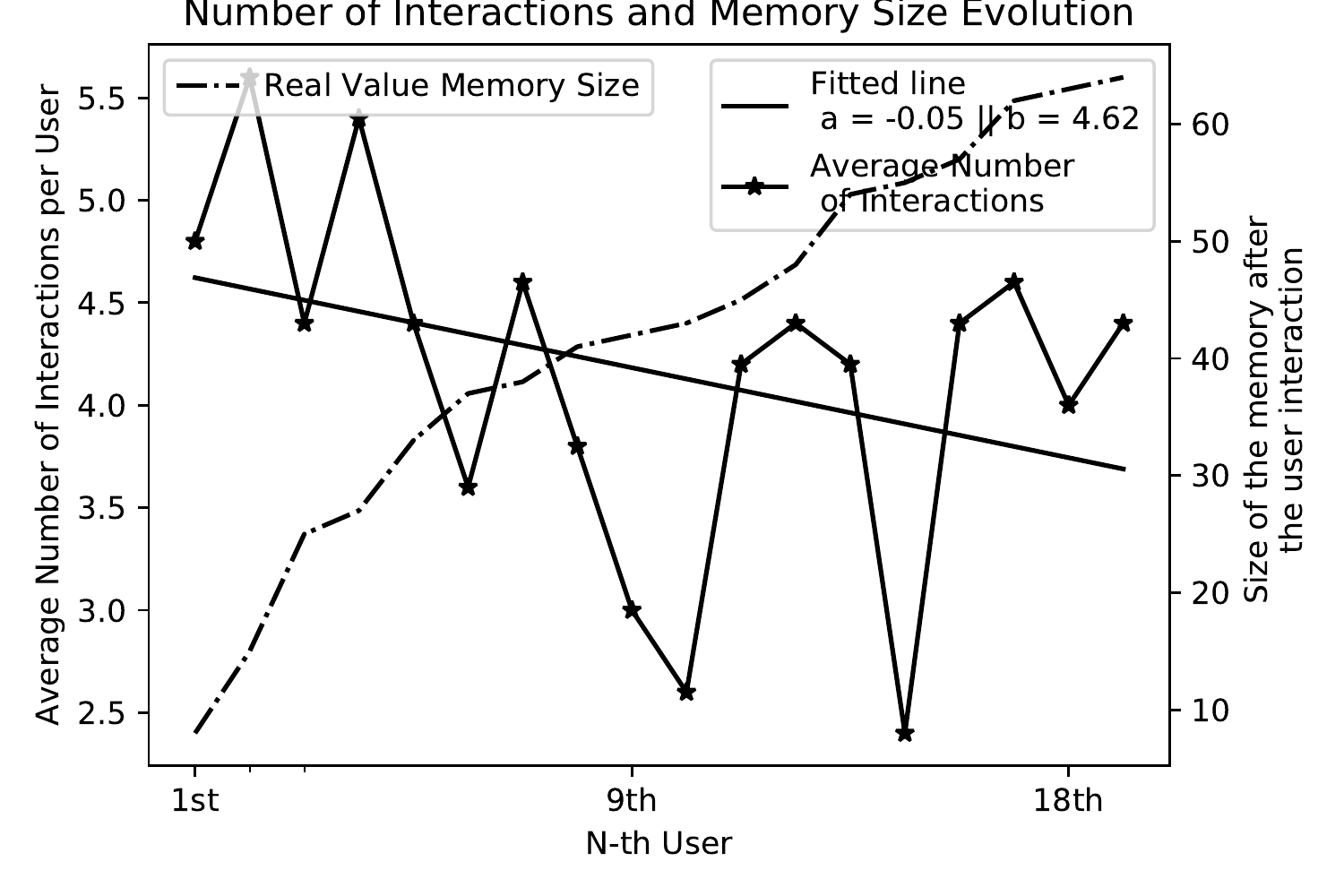}
\caption{NKB Learning from user interactions. DocuBot learns vocabulary through interaction with the several users. The number of interactions to create (and eventually clarify user's intent) decreases due to learning across users. }
\label{fig:experiment}
\end{figure}
In order to evaluate the NKB's learning ability, we tasked 18 users to interact with DocuBot for creating 5 slides using natural language commands. As discussed before, the NKB learns a new word $w_c$ it encounters by permanently mapping it to a sub-concept $\sigma$ and does not have the ability to forget the mapping. NKB assumes a perfect world scenario, where only collaborative users interact with DocuBot. However, as can be expected, in the real world wrong labels can exist. 

Figure \ref{fig:experiment} shows user interactions with DocuBot as well as NKB's growing memory size. In a robust setting, the expectation is that as the KB learns better representations and the memory grows, the $k$\textsuperscript{th} user does not have to enter as many commands as the ($k-1$)\textsuperscript{th} user. However as the figure indicates, the number of user interactions remains volatile with an average of $45\%$ decrease as the memory size consistently increases. Due to the limitations of the NKB, we use the RKB, which has the capacity to forget and re-learn proper representations.



\subsection{Robust Knowledge Base (RKB)}
In order to conduct a more extensive experiment with RKB, we simulate artificial users. This helps us to test RKB's performance over 1000+ users with many varying parameters. It also allows us to conduct comparative studies between RKB and NKB with identical usage patterns. 

We represent the set of main-concepts $\mu$ and sub-concepts $\sigma$ in our experiment using a dictionary. We also introduce \mbox{MatchingScore} as an evaluation metric to compare performance of both KB's.

We then simulate identical users in exactly the same order to compare the performance between the RKB and NKB using the \mbox{MatchingScore} metric, as well as demonstrate the superior performance of the RKB. We define it as:

\begin{equation}
     \mbox{MatchingScore(}V_{\hat{\sigma}}, V_{\zeta}\mbox{)} = \sum_{i=1}^{n}{\mathbbm{1}{(v_{\hat{\sigma_i}} = v_{\zeta_i})}}
\end{equation}
where $V_{\hat{\sigma}}$ is the vector of predicted sub-concepts ($\hat{\sigma}$) by the KB, and $V_{\zeta}$ is the vector of true (gold) sub-concepts. Note that $\norm{V_{\hat{\sigma}}}=\norm{V_{\zeta}}=\norm{V_{w_c}}$, where $V_{w_c}$ is the vector of words given by the user to create slides, .


\subsubsection{User Simulation}
To simulate a user, we generate a corpus of potential words humans would commonly use when interacting with DocuBot. We use Gensim \citep{rehurek_lrec} for loading the word vectors trained on the Google News dataset \citep{word2vec}. The model contains 300-dimensional vectors for 3 million words and phrases. The phrases were obtained using a simple data-driven approach described in \citep{Mikolov:2013:DRW:2999792.2999959}. Given a sub-concept $\sigma$, we use Word2Vec to retrieve the ordered list $L$ of the $N$ closest neighbors $w_c$. $N$ can be varied from small to large values to account for the diversity of vocabulary employed by a user. For each main-concept $\mu$, the simulated user selects a sub-concept $\sigma$ at random, then it picks with respect to a given \emph{pdf} a corresponding word $w_c$. The \emph{pdf} is chosen from a pool of distributions modeled as $\frac{1}{log(n)}$, $\frac{1}{n}$ or $\frac{1}{n^2}$, with $n \in \left[0,N\right]$. These models encapsulate the diverse behaviours from a wider to a more targeted vocabulary.

Learning from the NKB experimental results in \cref{fig:experiment}, we identify two types of users for simulation: (a) Collaborative users and (b) Non-Collaborative users. Users belonging to category (a) will use a $w_c$ belonging to the list $L$ of the correct sub-concept $\sigma$. Users belonging to category (b) will use a $w_c$ belonging to the list $L$ of another sub-concept. We define $\alpha$ as the ratio of Collaborative users to the total number of users in our experiments. In the experiments reported in this paper, we vary the threshold between 0.4 and 1,  
assuming in a real-world scenario at least $40\%$ of users are collaborative. We also vary the parameter $N$ between 5 and 1000, assuming that any user would know at least 5 words. 


\begin{figure*}[!htb] 
  \centering
  \begin{multicols}{3}
  \includegraphics[scale=0.3]{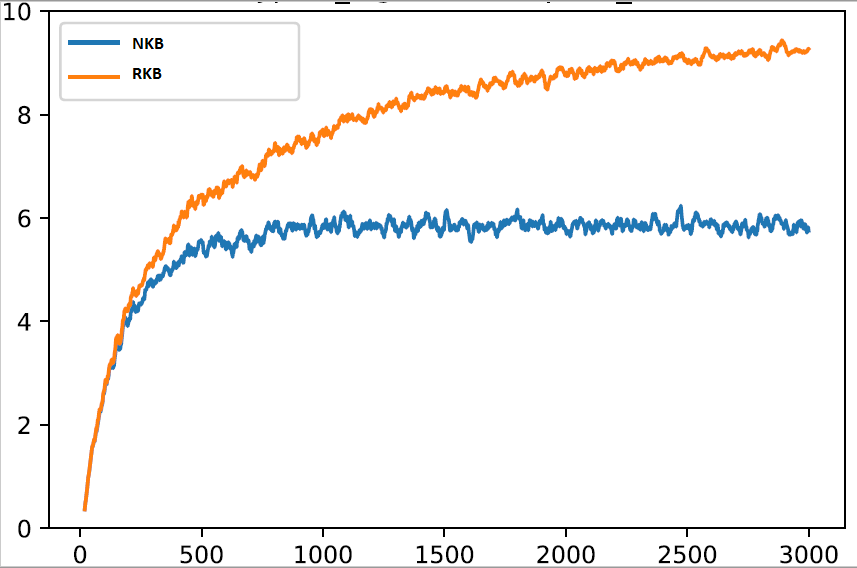}
  \subcaption{\emph{pdf}$\propto \frac{1}{log(n)}$}
  \par 
  \includegraphics[scale=0.3]{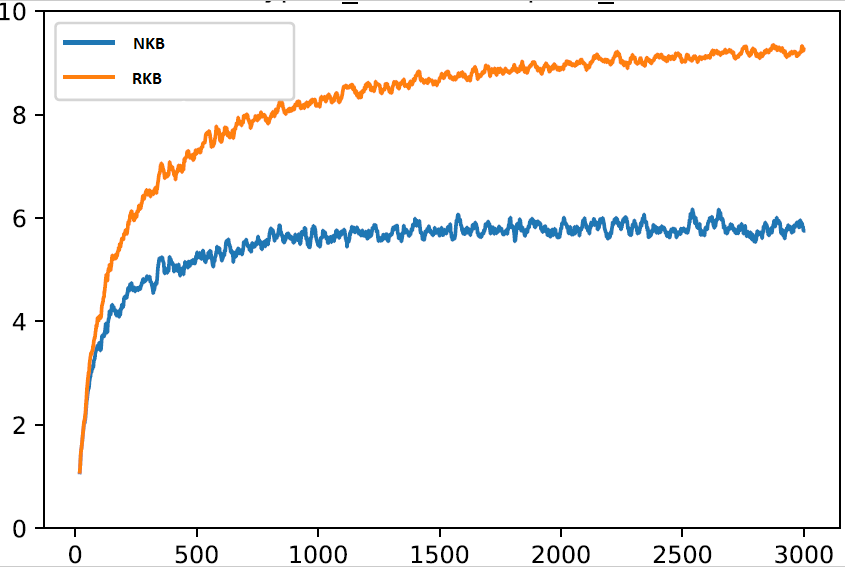}
  \subcaption{\emph{pdf}$\propto \frac{1}{n}$}
  \par 
  \includegraphics[scale=0.3]{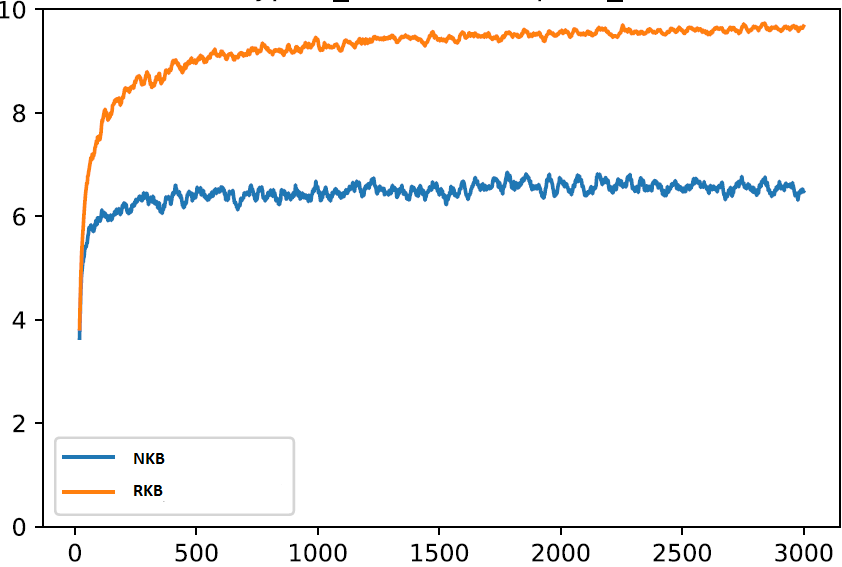}
  \subcaption{\emph{pdf}$\propto \frac{1}{n^2}$}
  \par 
  \end{multicols}

\caption{\emph{Learning phase} evolution of $MatchingScore$ in different experimental scenarios of the creation of 3000 simulated slides with $N=50$ and $\alpha = 0.6$ for different \emph{pdf} (Average of 10 simulations and smoothed over rolling window of 20). These experiments demonstrate the superior performance of Robust Knowledge Base (RKB), since it learns the new-vocabulary mappings faster, forgets incorrect mappings over-time resulting in a higher Matching Score compared to the Naive Knowledge Base (NKB). The smaller the variety of the users' vocabulary, the faster the KB learns.}
\label{fig:training1}
\end{figure*}

\begin{figure*}[!htb] 
  \centering
  \begin{multicols}{3}
  \includegraphics[width=1.2\linewidth]{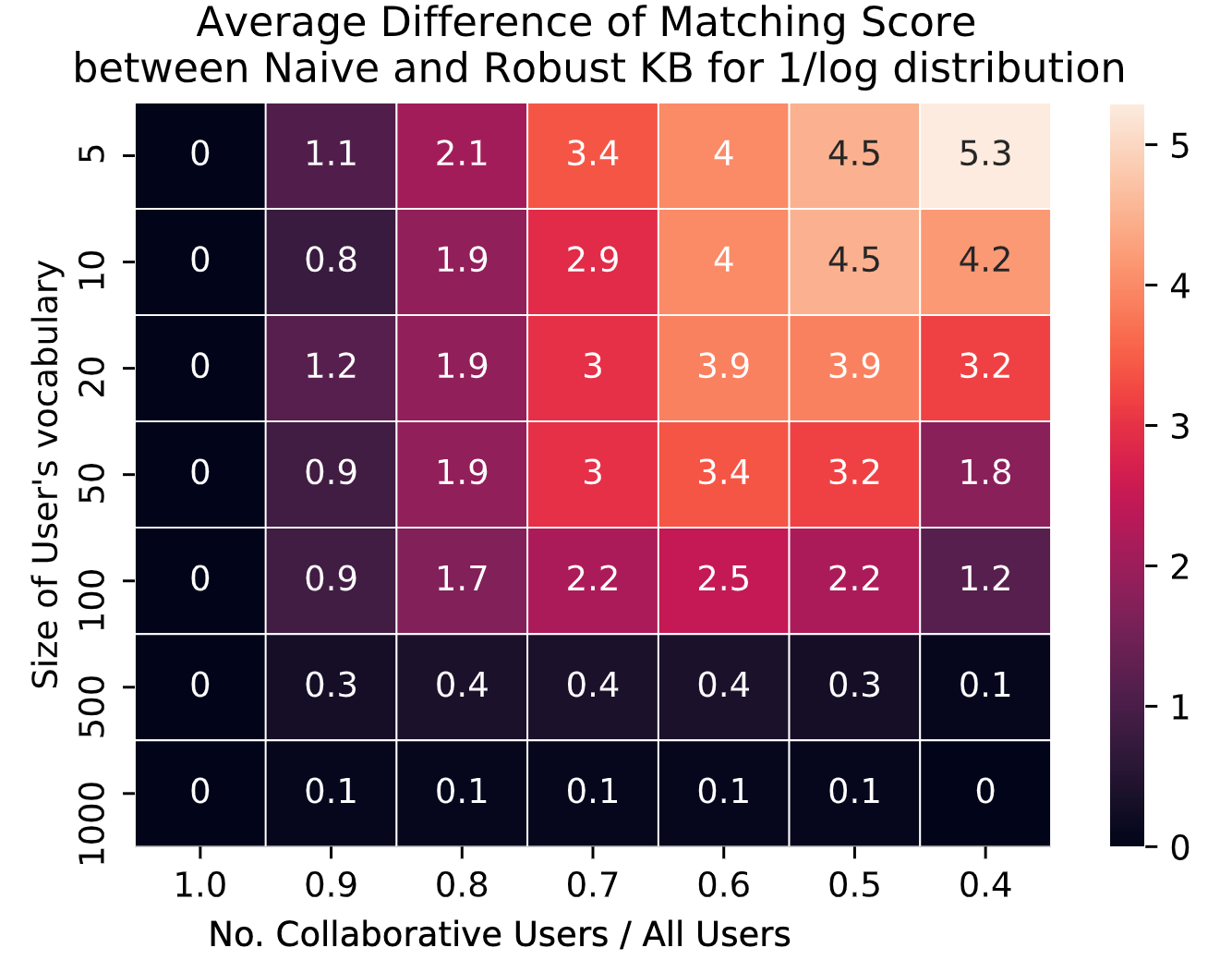}
  \subcaption{\emph{pdf}$\propto \frac{1}{log(n)}$}
  \par 
  \includegraphics[width=1.2\linewidth]{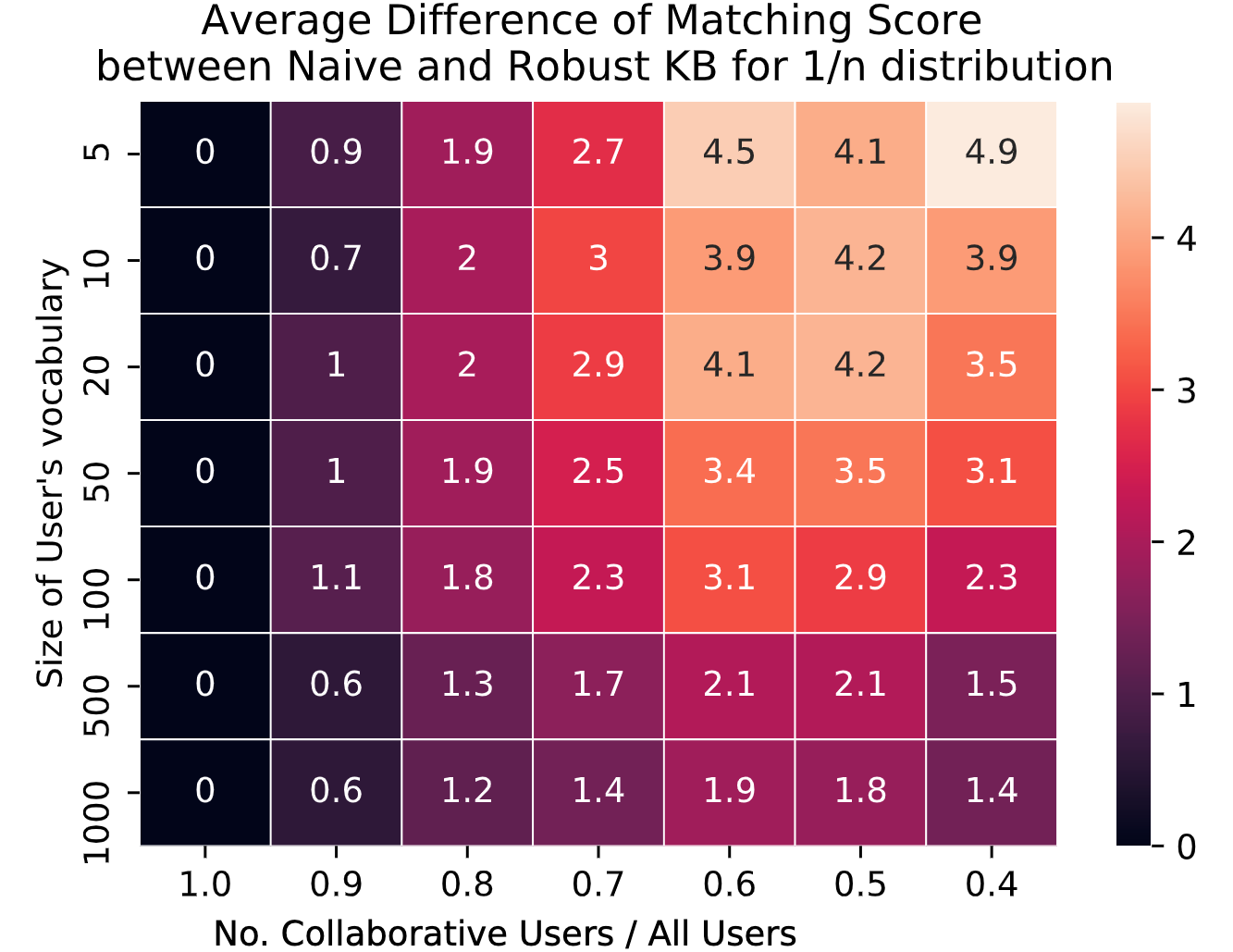}
  \subcaption{\emph{pdf}$\propto \frac{1}{n}$}
  \par 
 

    \includegraphics[width=1.2\linewidth]{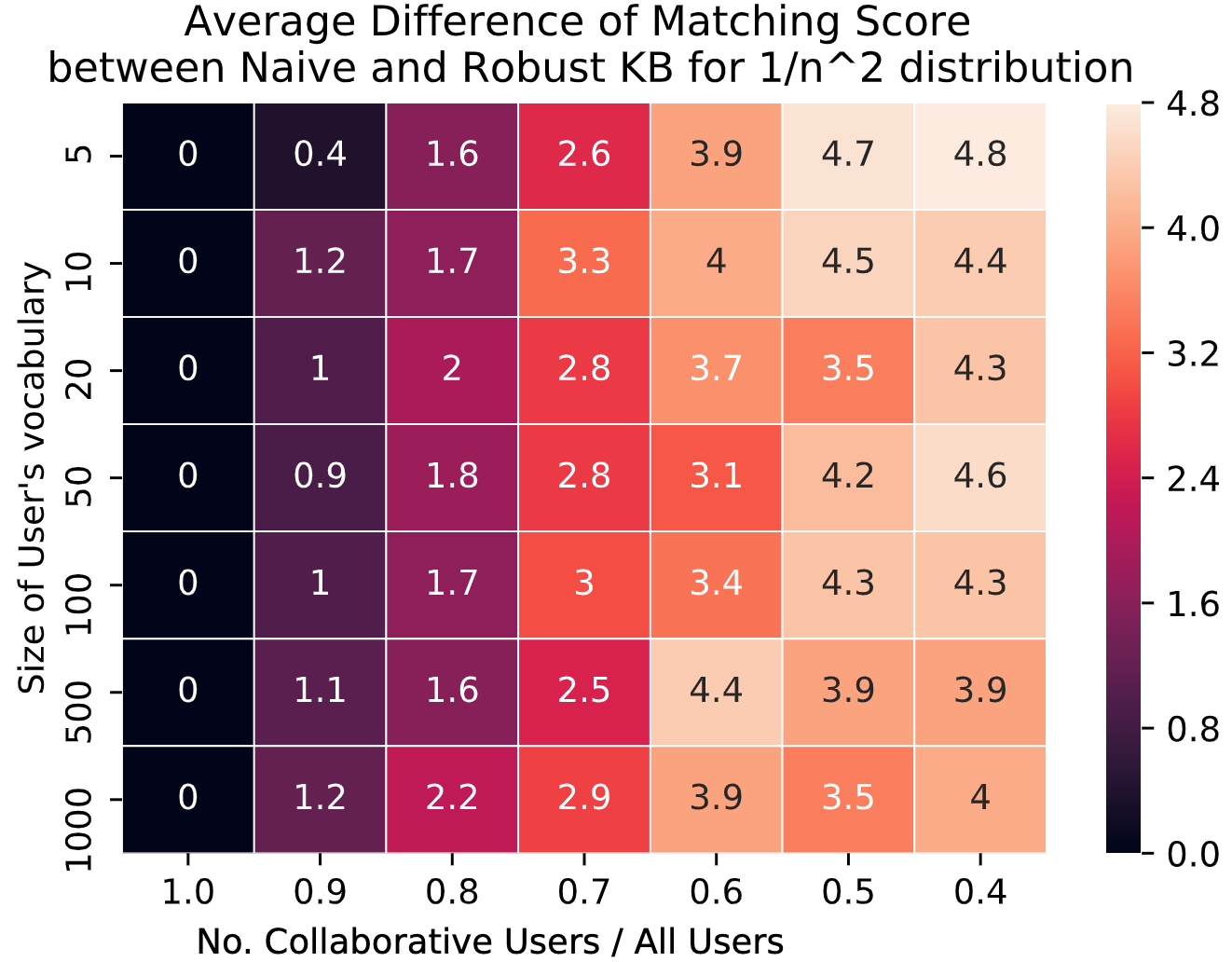}
  \subcaption{\emph{pdf}$\propto \frac{1}{n^2}$}
  \par
\end{multicols}

\caption{\emph{Evaluation phase} simulation results comparing NKB and RKB. Heatmap showing the average difference of $MatchingScore$ (score of RKB minus score of NKB) for the creation of 3000 slides. The $MatchingScore$ is between 0 and 10. The $x$ axis corresponds to the ratio $\alpha$ (ratio of Collaborative users to the total number of users), the $y$ axis corresponds to the vocabulary size of the user. These experiments show that the RKB consistently outperforms the NKB across all scenarios in terms of successful mapping between the user input and the text output in the document (higher Matching Score), in the presence of Non-Collaborative users.}
\label{fig:training2}
\end{figure*}

\subsubsection{Experiment Parameters}
The experiment is repeated E times, resetting the NKB and RKB every time. Each experiment is divided in two phases, a \emph{learning phase} and an \emph{evaluation phase}. During the learning phase, the two KBs are exposed to a proportion $\alpha$ $<$ 1 of Collaborative Users. In the evaluation phase, both KBs are exposed only to Collaborative Users. For each of the experiments (both in training and testing), the KBs are exposed to the creation of S slides. We average the results obtained across the E experiments and show them in Figures \ref{fig:training1} and \ref{fig:training2} (for these figures E=10 as it was a sufficient number of experiment to observe consistent results, and S=3000 as it was sufficient to observe stabilization in learning). Figure \ref{fig:training1} demonstrates that the Robust Knowledge Base (RKB) learns new vocabulary faster from simulated users compared to the Naive Knowledge Base (NKB). Figure \ref{fig:training2} demonstrates that the RKB consistently outperforms the NKB in all three scenarios having a higher Matching Score, with varying $\alpha$ (which controls the proportion of Non-Collaborative users).

\section{Illustrated Interface and Output}
\label{sec:userinterface}


DocuBot has been successfully deployed to generate different types of financial presentations and digital reports internally within J.P. Morgan. Investment Banking, Finance \& Business Management, Audit and External reporting teams etc. are some business use-cases, which have many applications for DocuBot. In this paper for purposes of demonstration, we illustrate an example of an user interacting with DocuBot to generate two PowerPoint slides\footnote{In practice, DocuBot can be used to create a large number of slides ($20+$) which is of great significance when creating Pitch-Books (a typical PowerPoint slide deck used for client communications) essential to Investment Banking teams.}.

\begin{figure*}
\begin{minipage}[c][10.4cm][t]{.5\textwidth}
  \vspace*{\fill}
  \centering
  \includegraphics[width=0.85\linewidth]{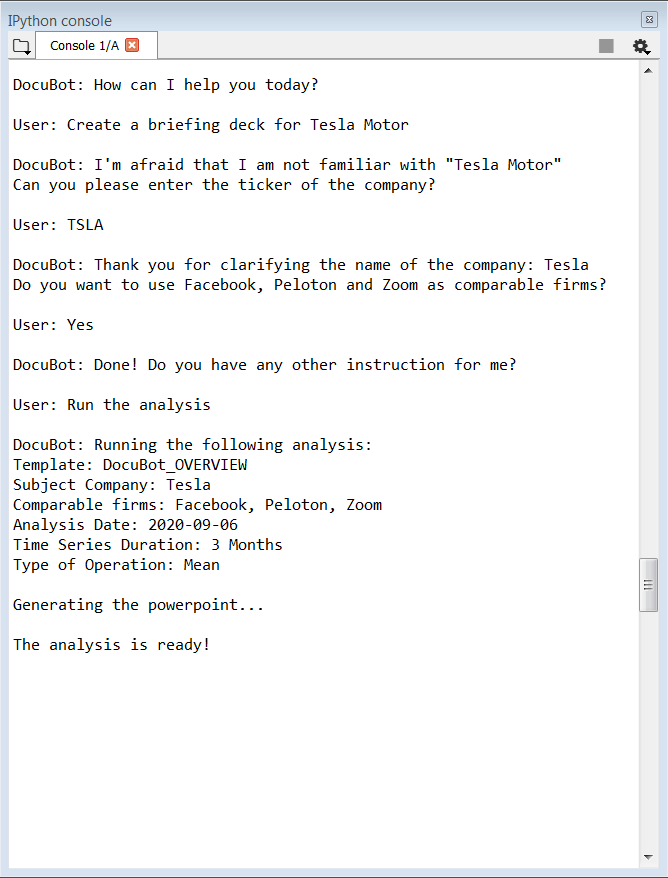}
  \subcaption{User - DocuBot interaction to create the deck of slides}
  \label{fig:interaction1}
\end{minipage}%
\begin{minipage}[c][10.4cm][t]{.5\textwidth}
  
  \centering
  \includegraphics[width=.95\linewidth]{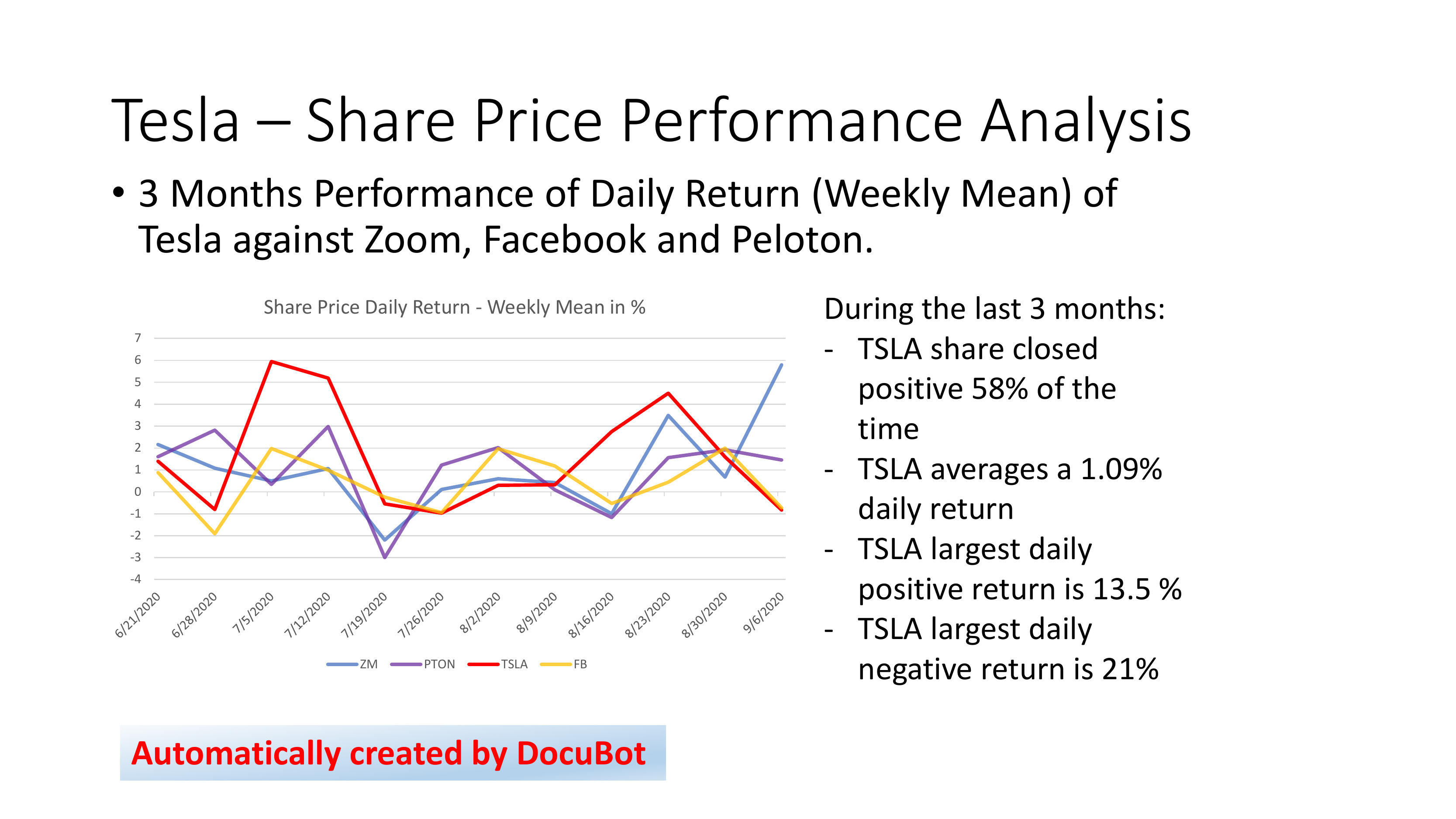}
  \subcaption{First page of the deck of slides}
  \label{fig:slide11}\par\vfill
  \includegraphics[width=.95\linewidth]{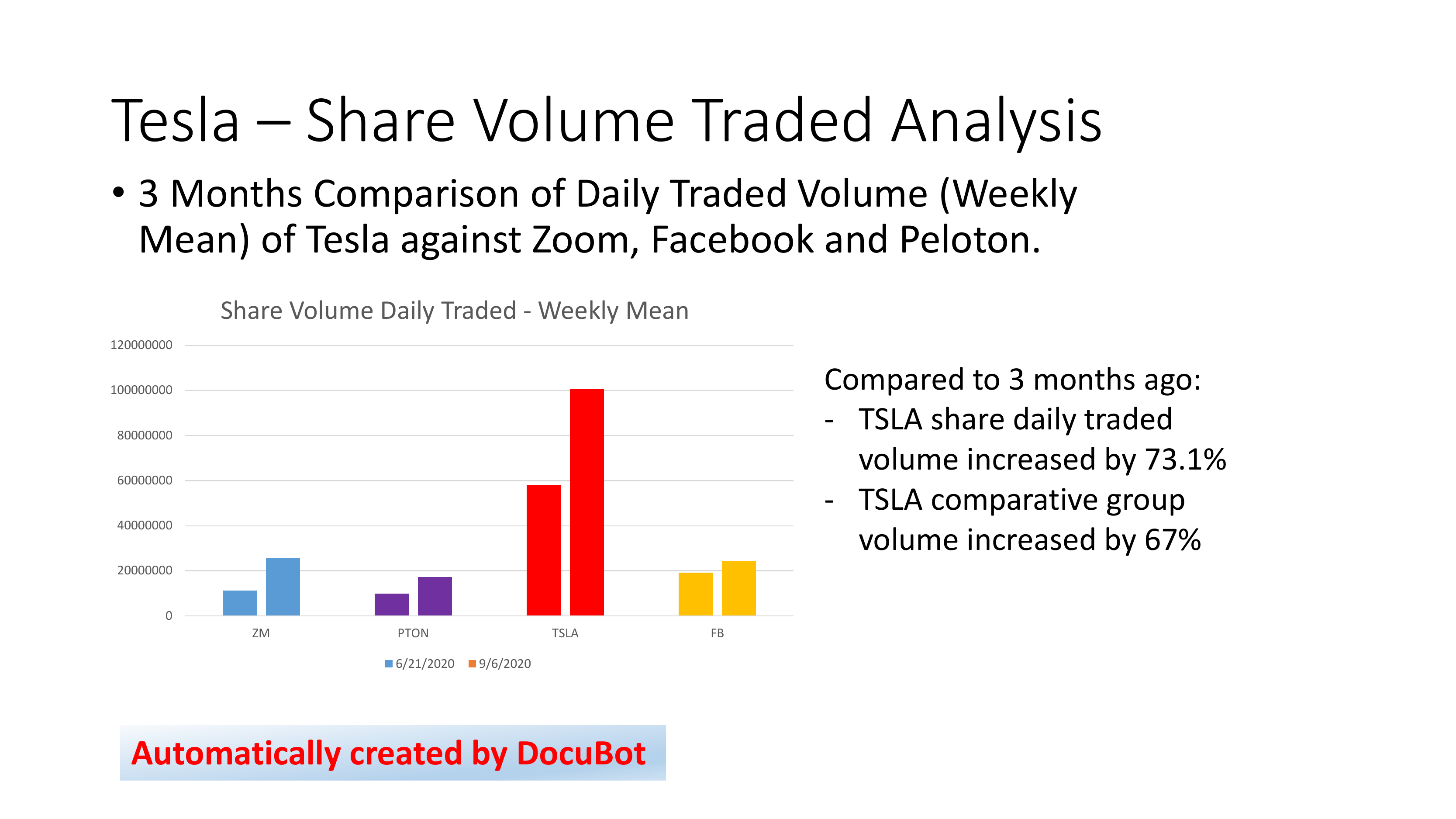}
  \subcaption{Second page of the deck of slides}
  \label{fig:slide12}
  \vspace*{\fill}
\end{minipage}
\caption{User provides a set of instructions and DocuBot creates the slide deck accordingly.}
\label{fig:create_slide}
\end{figure*}

\begin{figure*}
\begin{minipage}[c][10.5cm][t]{.5\textwidth}
  \vspace*{\fill}
  \centering
  \includegraphics[width=0.85\linewidth]{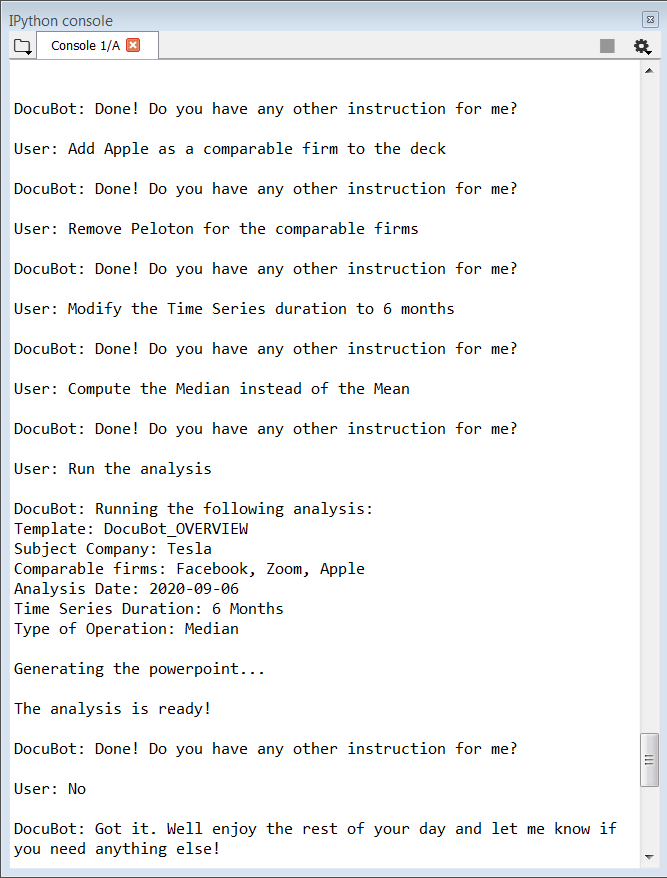}
  \subcaption{User - DocuBot interaction to modify the deck of slides}
  \label{fig:interaction2}
\end{minipage}%
\begin{minipage}[c][10.5cm][t]{.5\textwidth}
  
  \centering
  \includegraphics[width=.95\linewidth]{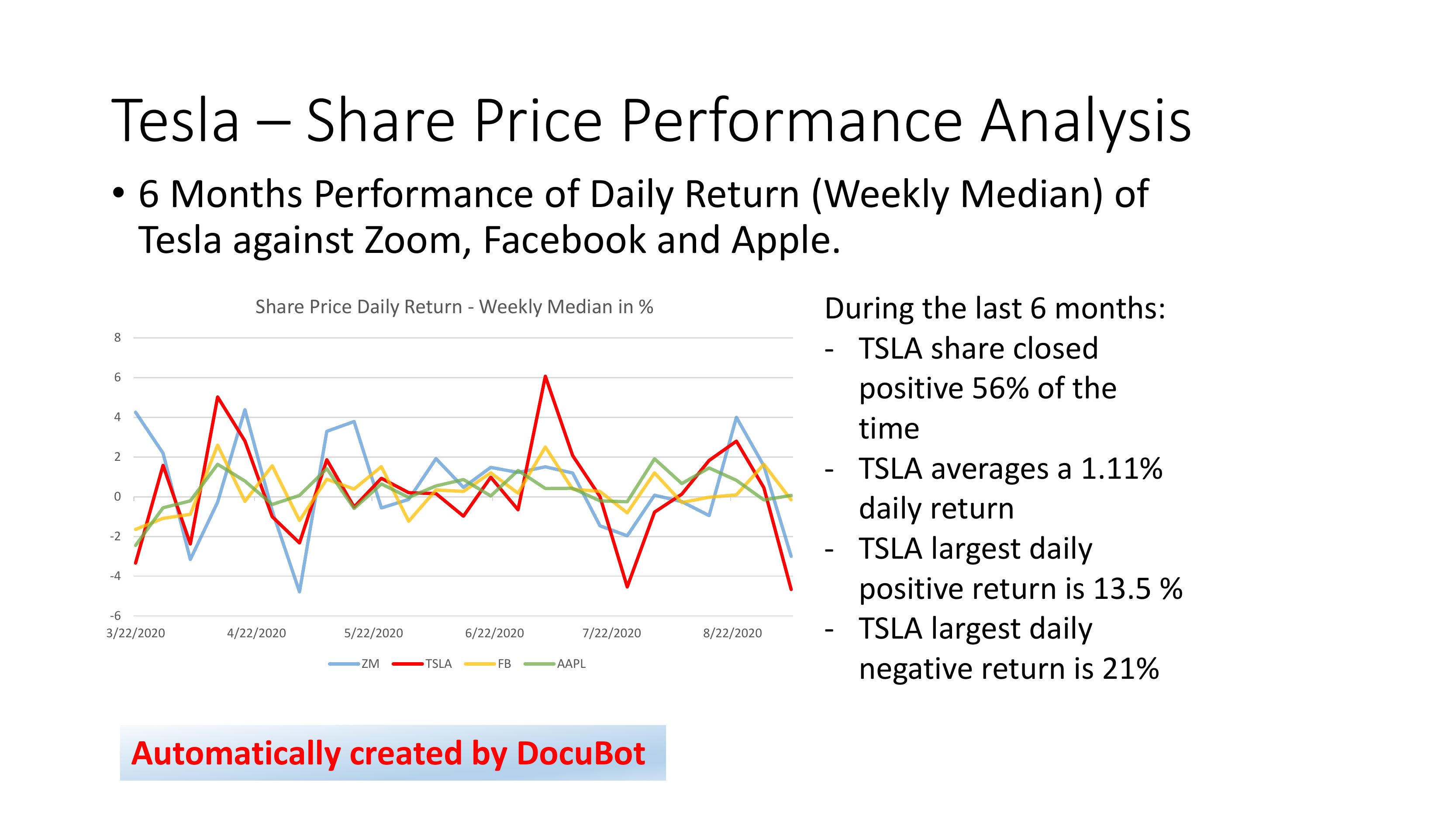}
  \subcaption{First page of the deck of slides}
  \label{fig:slide21}\par\vfill
  \includegraphics[width=.95\linewidth]{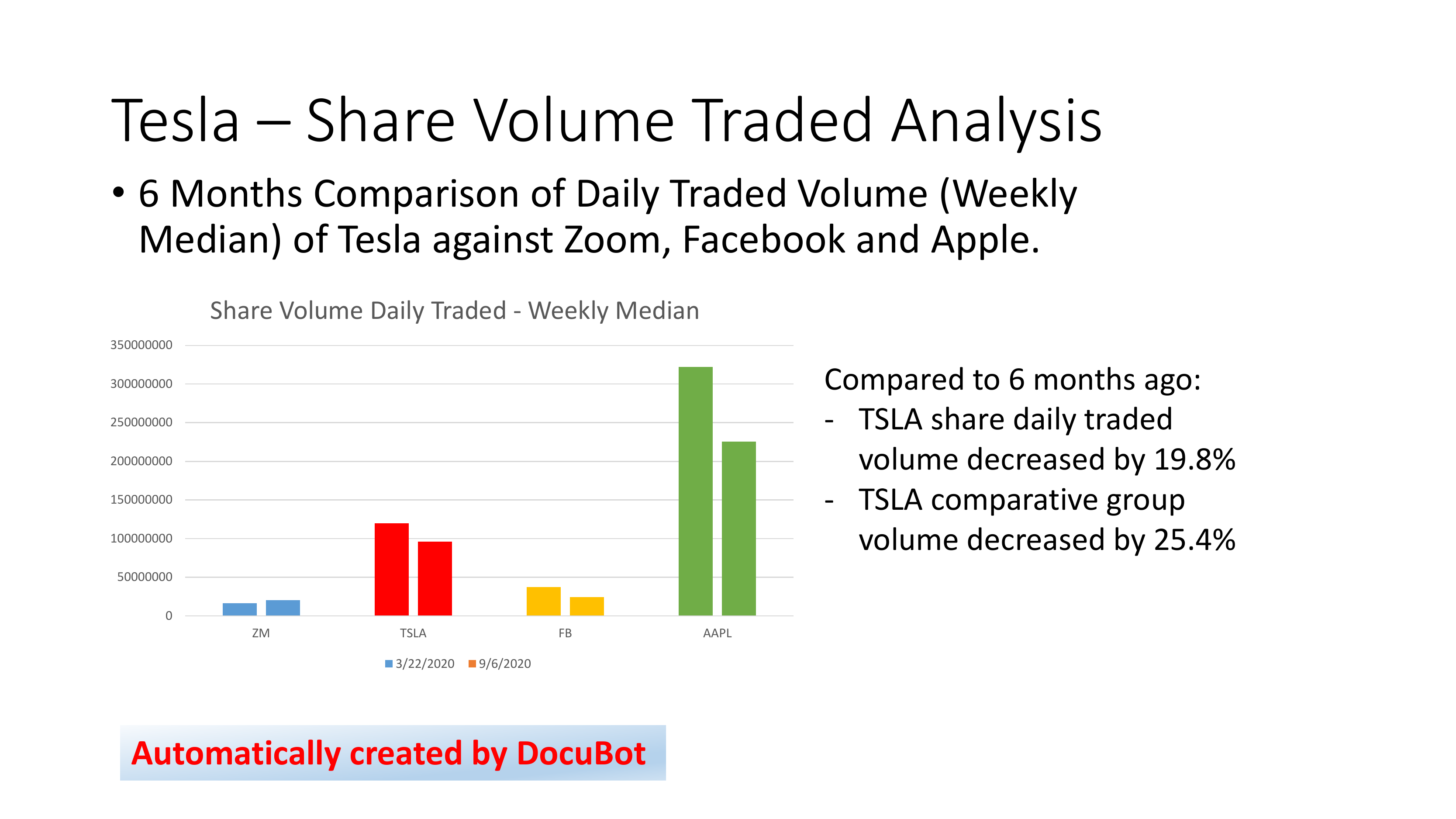}
  \subcaption{Second page of the deck of slides}
  \label{fig:slide22}
  \vspace*{\fill}
\end{minipage}
\vskip 0.2cm
\caption{User provides a second set of instructions and DocuBot modifies the slide deck accordingly.}
\label{fig:modify_slide}
\end{figure*}

The end-user is able to create and modify content in PowerPoint slides by interacting with DocuBot as illustrated in \cref{fig:interaction1} and \cref{fig:interaction2}. The user asks DocuBot to create a `briefing deck' about the company Tesla Motor. As seen in \cref{fig:slide11} and \cref{fig:slide12}, DocuBot ingests the OHLCV data \citep{yahoo} as input and creates two slides based on predefined templates ([`briefing deck']\textsubscript{object})  about the Share Price Performance and Share Volume Traded Analysis by interacting with the user. DocuBot's main features that are demonstrated in this interaction are listed below:-
\begin{itemize}
    \item The user asks DocuBot to create a `briefing deck' about `Tesla Motor'. However, this is not the official name of the company as saved in DocuBot's knowledge base, hence DocuBot cannot recognize it initially.
    \item DocuBot asks the user for clarification by specifying the ticker (unique name by which a company is publicly traded in financial markets) of the company. The ticker is used as a common key between the different words in the knowledge base. DocuBot can link `Tesla Motor' to the ticker `TSLA', linking to the company name `Tesla'.
    \item Docubot also has a list of commonly used comparable firms in its parameter configuration. This is a list of firms that are most frequently compared by a user to measure financial performance.
    \item User inputs the key sentence `Run the analysis', which launches the analysis using the default set of parameters.
\end{itemize}

We further demonstrate DocuBot's content modification capabilities in \cref{fig:interaction2}, where the user instructs DocuBot to modify the content of the slides in \cref{fig:slide11} and \cref{fig:slide12}. DocuBot is tasked with modifying parameters such as the list of comparable firms, horizon of analysis and metric for performance analysis:
\begin{itemize}
    \item The user asks to change the comparable firms by adding the company `Apple' and removing the company `Peloton'. The new comparable list of firms is automatically saved by DocuBot as a parameter configuration for future use by this user.
    \item The user asks to change time horizon of analysis from 3 months to 6 months. The displayed analysis now takes into account the last 6 months of time-series data instead of 3 months.
    \item The user asks DocuBot to use the `Median' instead of `Mean' as the metric for performance analysis. The operation made on the data is now a weekly median instead of the mean.
    \item User inputs the command `Run the analysis', which re-generates the final output slides using the new set of parameters shown in \cref{fig:slide21} and \cref{fig:slide22}. 
\end{itemize}

Internally across the firm, we have extensively tested and evaluated DocuBot to generate several business reports and presentations successfully.

\section{Contributions and Impact}
DocuBot is a use-proven tool at J.P. Morgan that contributes in several directions: (1) DocuBot leverages its learning capability across users in order to learn the particular or domain-specific vocabulary employed by the groups of users. (2) The use of a Robust Knowledge Base enables DocuBot to correct its memory if affected by malicious users, this capacity to forget gives it the ability to adapt to the evolution of the vocabulary across time. (3) DocuBot is also capable of \emph{Saving and Reusing Skills} enabling the users to define their own \emph{Macro Skills} and eventually re-use them in order to save time. It becomes very easy to create new reports---the user only needs to create the new skills and add them to the \emph{skills library}. (4) DocuBot enables a \$0 cost \emph{what-if} scenario analysis, since it is very easy to modify the document parameters and to get the output instantaneously. (5) DocuBot automatically generates text from numerical data. This intelligent text generation is possible through the use of \emph{insights ranking and selection} in order to output only the most interesting facts. Internally at J.P. Morgan, feedback from users suggested that the automation introduced by DocuBot to help business analysts could potentially reduce their time spent in creating and updating PowerPoint slides from \textit{over 5 hours} to \textit{less than 1 minute}.



\section{Conclusion and Future Work}
In this paper, we have introduced a novel framework, DocuBot, to automate the generation of digital reports like PowerPoint slides through human-AI interaction. To provide an easily interpretable explanation of the data displayed on the slides, we have also introduced the automated generation of AI Insights in these presentations. We have demonstrated the robustness of DocuBot to adapt to different types of users through several experiments. As we continue to enhance several components of the framework, internally at J.P. Morgan, we have deployed DocuBot to demonstrate its capabilities in generating financial presentations for real world use cases. In addition, by applying the DocuBot framework internally as well as through public stock market data, we have demonstrated the broad applicability of this emerging technology across the industry.

We are currently exploring a few venues to enhance DocuBot's capabilities. We are experimenting with more flexible data-to-text generation methods involving Probabilistic Context-Free Grammars, when compared to the template-based approach presented here. We are also exploring methods to make it easier for DocuBot to scale to new use cases such as methods for accommodating new types of analysis, custom insight-scoring functions, and a source-agnostic data ingestion layer. \\

\footnotesize

\textbf{Acknowledgements}

We would like to thank Vamsi Alla, Jiahao Chen, Kwun Lo, and the Global Investment Banking team at J.P. Morgan for their contributions.\\

\textbf{Author Contributions}

The first two authors contributed equally to this work. \\

\textbf{Patent Pending}

There is a commercial interest in leveraging the use of this framework and a patent application has been filed. \\

\scriptsize

\textbf{Disclaimer}

This paper was prepared for informational purposes by the Artificial Intelligence Research group of JPMorgan Chase $\&$ Co and its affiliates (“J.P. Morgan”), and is not a product of the Research Department of J.P. Morgan.  J.P. Morgan makes no representation and warranty whatsoever and disclaims all liability, for the completeness, accuracy or reliability of the information contained herein.  This document is not intended as investment research or investment advice, or a recommendation, offer or solicitation for the purchase or sale of any security, financial instrument, financial product or service, or to be used in any way for evaluating the merits of participating in any transaction, and shall not constitute a solicitation under any jurisdiction or to any person, if such solicitation under such jurisdiction or to such person would be unlawful.

\normalsize

\bibliography{bib}

\end{document}